\documentclass[journal]{IEEEtran}
\usepackage{amssymb}
\usepackage{graphicx}
\usepackage{epstopdf}
\usepackage{enumerate}
\usepackage{amsmath}

\usepackage{amsmath}
\usepackage{multirow}

\usepackage{subfigure}

\usepackage[pagebackref=true,breaklinks=true,letterpaper=true,colorlinks,bookmarks=false]{hyperref}

\newcommand{\etal}{\textit{et al}.}
\newcommand{\ie}{\textit{i}.\textit{e}.,}
\newcommand{\eg}{\textit{e}.\textit{g}.}

\usepackage{url}
\urldef{\mailsa}\path|{alfred.hofmann, ursula.barth, ingrid.haas, frank.holzwarth,|
\urldef{\mailsb}\path|anna.kramer, leonie.kunz, christine.reiss, nicole.sator,|
\urldef{\mailsc}\path|erika.siebert-cole, peter.strasser, lncs}@springer.com|

\begin{document}
%
\title{Single Image Super-resolution via a Lightweight Residual Convolutional Neural Network}
%
%
%

\author{Yudong~Liang,
        Ze~Yang,
        Kai~Zhang,
        Yihui~He,
         Jinjun~Wang,~\IEEEmembership{Senior Member,~IEEE,}
        and~Nanning~Zheng,~\IEEEmembership{Fellow,~IEEE}
\thanks{Yudong Liang is with the Key Laboratory of Computational Intelligence and Chinese Information
Processing of Ministry of Education, and the School of Computer and
Information Technology, Shanxi University, 92 Wucheng Road, Taiyuan, Shanxi Province, 030006, China. A large part of the work is done when he was with the Institute of Artificial Intelligence and Robotics, Xi'an Jiaotong University, China. e-mail: liangyudong006@163.com.}
\thanks{Ze Yang, Yihui He, Jinjun Wang, and Nanning Zheng are with the Institute of Artificial Intelligence and Robotics, Xi'an Jiaotong University, China.}
\thanks{Kai~Zhang is with Harbin Institute of Technology, China}
}

\markboth{}%
{Shell \MakeLowercase{\textit{et al.}}: Bare Demo of IEEEtran.cls for IEEE Journals}
%



\maketitle

\begin{abstract}
Recent years have witnessed great success of convolutional neural network (CNN) for various problems both in low and high level visions. Especially noteworthy is the residual network which was originally proposed to handle high-level vision problems and enjoys several merits. This paper aims to extend the merits of residual network, such as skip connection induced fast training, for a typical low-level vision problem, i.e., single image super-resolution. In general, the two main challenges of existing deep CNN for supper-resolution lie in the gradient exploding/vanishing problem and large numbers of parameters or computational cost as CNN goes deeper. Correspondingly, the skip connections or identity mapping shortcuts are utilized to avoid gradient exploding/vanishing problem. In addition, the skip connections have naturally centered the activation which led to better performance.
To tackle with the second problem, a lightweight CNN architecture which has carefully designed width, depth and skip connections was proposed. In particular, a strategy of gradually varying the shape of network has been proposed for residual network. Different residual architectures for image super-resolution have also been compared. Experimental results have demonstrated that the proposed CNN model can not only achieve state-of-the-art PSNR and SSIM results for single image super-resolution but also produce visually pleasant results. This paper has extended the mmm 2017 oral conference paper with a considerable new analyses and more experiments especially from the perspective of centering activations and ensemble behaviors of residual network.
\end{abstract}

\begin{IEEEkeywords}
super-resolution, deep residual convolutional neural network, skip connections,  parameter numbers
\end{IEEEkeywords}

%
\IEEEpeerreviewmaketitle

\section{Introduction}
%
%
%
%
\IEEEPARstart{S}{ingle} image super-resolution (SISR)~\cite{zhang2015revisiting,zhang2016joint,liang2015incorporating} aims to recover a high-resolution (HR) image from the corresponding low-resolution (LR) image. It is a very practical technique due to its high value in various fields, such as producing high-definition images from low-cost image sensors, medical imaging and satellite imaging. Restoring the HR image from the single LR input is also a very difficult problem of high theoretical values which arouses more and more interests from the academic communities~\cite{timofte2016seven,dong2016image,yang2014singleBenchmark,huang2015single} and large companies~\cite{romano2017raisr,ledig2017photo,shi2016real}. Typically, it is very challenging to restore the missing pixels from an LR observation since the number of pixels to be estimated in the HR image is usually much larger than that in the given LR input. The ill-pose nature of single image super-resolution problem makes restoring HR images an arena to evaluate inference and regression techniques. Generally, SISR techniques can be roughly divided into three categories: the interpolation methods, the reconstruction methods~\cite{irani1993motion} and the example based methods~\cite{freeman2000learning,yang2008image}.

Most of the recent SISR methods fall into the example based methods which learn prior knowledge from LR and HR pairs, thus alleviating the ill-posedness of SISR. Representative methods mainly rely on different learning techniques to incorporate image priors for super-resolution process, including neighbor embedding regression~\cite{chang2004super,timofte2013anchored,timofte2014a+}, sparse coding~\cite{yang2008image,yang2010imageTip}, tree based regressions~\cite{schulter2015fast,Salvador2015} and deep convolutional neural network (CNN)~\cite{dong2014learning,dong2016image,liang2016incorporating,kim2016accurate}.

\begin{figure*}[th]
     \centering
     \footnotesize
     \includegraphics[width=0.850\textwidth]{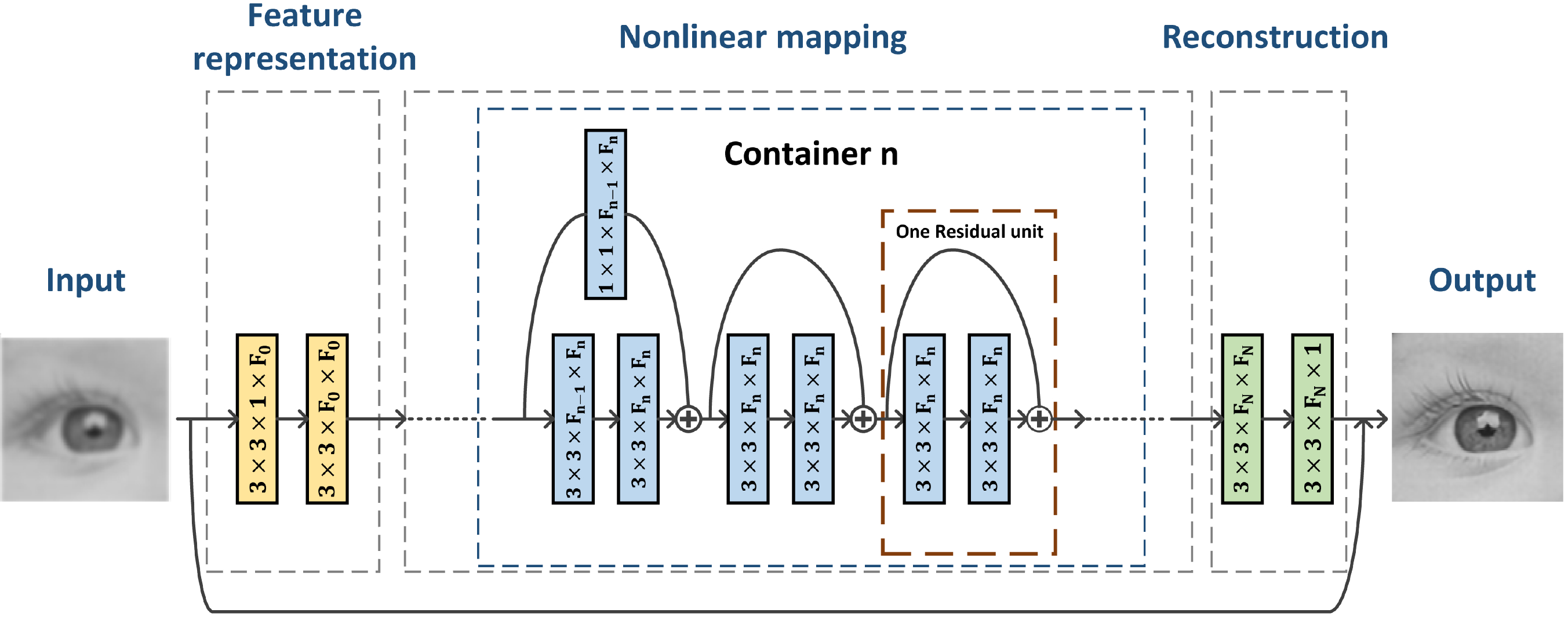}
     \caption{The architecture of our residual model.}
  \label{fig:arch}
\end{figure*}

Among the above techniques, deep learning techniques especially deep CNN have largely promoted the state-of-the-art performances in SISR area. Dong \etal \cite{dong2014learning} firstly proposed a deep convolutional neural network termed SRCNN with three convolutional layers for image super-resolution which gave the best practise at that time.
Later, Dong \etal \cite{dong2016image} extended SRCNN with larger filter sizes and filter numbers while kept the depth of CNN fixed to further improve the performance. They found that deeper models were hard to train and \cite{dong2016image} failed to boost the performance by increasing the depth. Such findings indicate that deeper models are not suitable for image super-resolution, which is counter-intuitive as deeper models have been proved more effective in many tasks~\cite{simonyan2014very,he2015deep,he2016identity}. Instead of directly predicting the HR output,
Kim \etal \cite{kim2016accurate} proposed a very deep CNN (VDSR) of depth up to 20 by a large skip connection to predict the residual image, \ie the high frequency of the HR image. VDSR surpasses SRCNN with a large margin which mainly benefits from two aspects: deeper architecture and predicting high frequency of images only which is called residual learning by \cite{kim2016accurate}.  

As demonstrated in~\cite{kim2016accurate}, the SR results have been improved as the VDSR network goes deeper to a certain depth (20). Although VDSR has achieved impressive results, the plain structure of VDSR which simply stacks layers hampers the convergence of deeper architectures due to the gradient exploding/vanishing problem. It would bring little improvement as the network goes deeper. Fortunately, the residual network~\cite{he2015deep,he2016identity} has successfully addressed this issue. As a result, different from VDSR, this paper has designed a novel very deep residual convolutional neural network shown in Fig.~\ref{fig:arch}. As LR image and target HR image are highly correlated, predicting high frequency of the image only is a kind of residual learning which largely lowers the price for training.
A totally deep residual CNN will be expected to fully take advantage of the correlations between LR and HR images. Moreover, skip connections or identity mapping shortcuts in deep residual CNN would alleviate gradient vanishing/exploding problem when the network becomes increasingly deeper. 

For neural network it is known that mean shifts toward zero or centering the activations speeds up learning~\cite{le1991eigenvalues,orr2003neural} by bringing the normal gradient closer to the unit natural gradient~\cite{amari1998natural}. LeCun \etal~\cite{le1991eigenvalues} justified the benefits of centering activation functions and advised to center the distribution of supervision information. The mean value of distribution for high frequency in the images is around zeros, while the mean of raw HR image pixels biases above zero. Thus, it is easy to understand that predicting residual images (high frequency) instead of HR images has largely improve the convergence of the network~\cite{kim2016accurate}. Batch Normalization (BN)~\cite{ioffe2015batch}
also aimed to center activations by reducing the internal covariate shift with extra moving average computations.
Raiko \etal~\cite{raiko2012deep} proved that shortcut connections made the Fisher information matrix closer to a diagonal matrix and standard gradient closer to the natural gradient which contributes to centering the outputs and slopes of the hidden layers in multi-layer perceptron network. Thus, identity mapping shortcuts naturally help the residual network center the activations.

The Batch Normalization (BN) layers in the conventional residual branches~\cite{he2015deep,he2016identity} are abandoned in our proposed architecture as skip connections and predicting high frequency (with a zero mean) have ensured centering the activations if the network is not too deep. 
Our designed residual architectures which we refer to as ``SRResNetNB'' have consumed less computational resources and achieved better performances empirically.

While very deep CNN model would increase the model capacity, on the other hand, it would introduce a huge number of parameters which is sometimes unacceptable for applications. Thus, when hardware resources are limited, a lightweight architecture using less parameters is essential for real word applications. In this paper, the `shape' of deep CNN has been investigated to largely reduce the parameter numbers. The `shape' of deep CNN refers to depth, all the filter sizes and numbers of each layer which decide sizes and numbers of feature maps in each layer to form a global shape.  
With a residual architecture and lightweight `shape' design, the proposed model can not only achieve state-of-the-art PSNR and SSIM results for single image super-resolution but also produce visually pleasant results.

A preliminary version of this work was presented earlier~\cite{yang2017single}. The present work adds to the conference oral version in significant ways: first, different deep architectures of residual branches are explored to further conclude a principle of designing a deep network for image super-resolution. Second, a considerable new
analysis from the perspective of centering activations and ensemble behaviors of residual networks has been represented and intuitive explanations are supplied to the result. In particular, a strategy of gradually varying the `shape' of the residual network has been clarified in constructing a lightweight structure, based on the assumption that the residual network has been seen as an ensemble of relatively shallow networks with a large capacity~\cite{Veit2016Residual}. Third, more detailed experiments are represented to design the structures and retrench the parameters of the residual model.

\section{Related Works}

In the pioneer work by Freeman~\etal~\cite{freeman2000learning}, the co-occurrence priors were proposed that similar LR local structures often relate to similar HR local information. From LR and corresponding HR images, LR and HR examples (patches or sub images) could be extracted to form training databases. The mappings from LR to HR examples call for accurate regression methods to be applied. In fact, the learning based regression methods especially deep learning based methods have dominated the example based methods.

%

Since the work of SRCNN~\cite{dong2014learning}, deep CNNs have refreshed the state-of-the-art performances in super-resolution area. Simply elaborating the filter sizes and filter numbers for SRCNN~\cite{dong2016image} had further improved the performance. Wang~\etal~\cite{wang2015deep} designed the CNN architecture to incorporate the sparse coding prior based on the learned iterative shrinkage and thresholding algorithm (LISTA). With sparsity prior modeling, the performance boosted even with a model of smaller size compared with SRCNN.

Kim~\etal~\cite{kim2016accurate} made a breakthrough for image super-resolution by predicting residual images and using a much deeper CNN termed VDSR up to 20 layers, which largely accelerated the speed of training and outperformed SRCNN presented by Dong~\etal~\cite{dong2016image}. To ensure the fast convergence of deep CNN and avoid gradient vanishing or exploding, a much larger learning rate for training was cooperated with adjustable gradient clipping in VDSR training. VDSR is inspired by the merits of VGG net which attempts to train a thin deep network. However, this kind of plain networks are not easy to be optimized when they go even deeper as demonstrated by He~\etal~\cite{he2015deep,he2016identity}.

The difficulties of training deeper plain networks were carefully analyzed by He~\etal~\cite{he2015deep,he2016identity}. The degradation problem~\cite{he2015deep} has been observed that the testing accuracy even the training accuracy becomes saturated then degrades rapidly as plain networks go deeper. This degradation is caused by the difficulties of training other than overfitting. It has been demonstrated that learning a residual function is much more easier than learning the original prediction function with very deep CNN. Residual networks with a surprising depth were designed for image classification problems with skip connections or identity mapping shortcuts. Later, a detailed analysis on the mechanisms of identity mapping in deep residual networks and a new residual unit design have been represented in~\cite{he2016identity}.

Residual network has also been applied in conjunction with perceptual loss~\cite{johnson2016perceptual} to generate style transferred image and produce visual more pleasing HR images in large magnification factors. SRResNet~\cite{ledig2017photo}, another famous concurrent work with us has also designed a residual network with skip connections for image super-resolution which serves as the generator network in a generative adversarial framework, termed SRGAN. To produce photo-realistic results, SRGAN~\cite{ledig2017photo} exploited an adversarial loss to replace the traditional mean squared reconstruction error. This adversarial framework recovered images with better perceptual quality and especially favored large upscaling factors (\eg, 4). The success of these work and our previous version~\cite{yang2017single} has indicated the importance of skip-connections for image super-resolution. Later, Tai \etal~\cite{tai2017image} introduced skip connections of multiple paths and shared the weights of residual units in one recursive block. Most of the residual networks for image super-resolution designed the residual branches as a combination of convolution, nonlinear activation (such as ReLU or PReLU) and Batch Normalization (BN) layers, which are the same as the residual branches for image classification task.

The idea of shortcuts has been related with centering the activations at zero for multi-layer perceptron network~\cite{raiko2012deep}. Raiko \etal~\cite{raiko2012deep} proposed to transform the outputs of each hidden layers in multi-layer perceptron network to have zero output and zero slope on average and use separate shortcut connections to model the linear dependencies. It is known that centering the activations accelerates learning~\cite{clevert2015fast,raiko2012deep,le1991eigenvalues,orr2003neural}. LeCun \etal~\cite{le1991eigenvalues} analyzed the eigenvalues of Hessian matrix during the gradient descend process and give a theoretical justification for the benefits of centering activation functions.
The applied skip-connections have already centered the activations at zero within certain depth and the mean of distributions for high frequencies in images is close to zero, which indicate the BN layers could be eliminated in our residual units.

Our design has been further supported by the very recent work~\cite{lim2017enhanced}, which wons the first prize in Ntire 2017 challenge on single image super-resolution~\cite{timofte2017ntire}. Liang \etal~\cite{liang2017single} further extended the identity skip connections to projection skip connections and explored the power of internal priors for deep architectures.

After largely easing the difficulties of training much deeper CNN with residual functions by shortcuts or skip connections, the huge number of parameters in deep architecture is still a big problem for computational resources and storages. The evolvement of Inception models~\cite{szegedy2015going,ioffe2015batch,szegedy2016rethinking,szegedy2017inception} has demonstrated that carefully designed topologies enable compelling performances with less parameters. He \etal~\cite{he2015deep,he2016identity} attempt to alleviate the problem by bottleneck architectures.
The bottleneck architectures first utilize $1\times1$ convolutions to reduce the dimensions, then after some operations, $1\times1$ convolutions are applied again to increase the dimensions. With such a residual unit design, the number of parameters could be largely reduced. Thus, the `shape' of CNN could be potentially explored to reduce the parameters while maintain the performances. The bottleneck architectures decrease the parameter numbers at the expense of increasing the depth of the network to mountain the performances. In the meanwhile, contextual information is very important for image super-resolution~\cite{kim2016accurate,dong2016image}, such $1\times1$ convolutions design may give a negative effort to the SR results. 
A study on the influences of the `shape' (the filter sizes, depth and numbers of convolutions in each layer) on the performance of image super-resolution has been represented in the following sessions.
With a carefully design and exploration of the `shape' of the network, novel residual deep models are proposed for image super-resolution task in this paper.

\section{A lightweight Residual Deep Model for Image Super-resolution}

Following the example based methods, HR examples $I^h$ and LR examples $I^l$ are extracted from HR images $I^H$ and LR images $I^L$ respectively.
The degeneration process of LR images $I^L$ from the corresponding HR images $I^H$ could be considered as the following blurring process related with blur kernel $G$ and downsampling process $\downarrow_{s}$ with a a scale factor $s$ 
            \begin{equation}\label{eq:downsample}
                 I^L=(I^H\otimes{G})\downarrow_{s}.
             \end{equation}
In the experiments, this process is simulated by a `bicubic' downscale interpolation.

In the next part, residual deep models for image super-resolution will be designed from the perspective of centering the activations to speed up learning.

\begin{figure*}[!htb]
     \centering
     \footnotesize
     \includegraphics[width=0.750\textwidth]{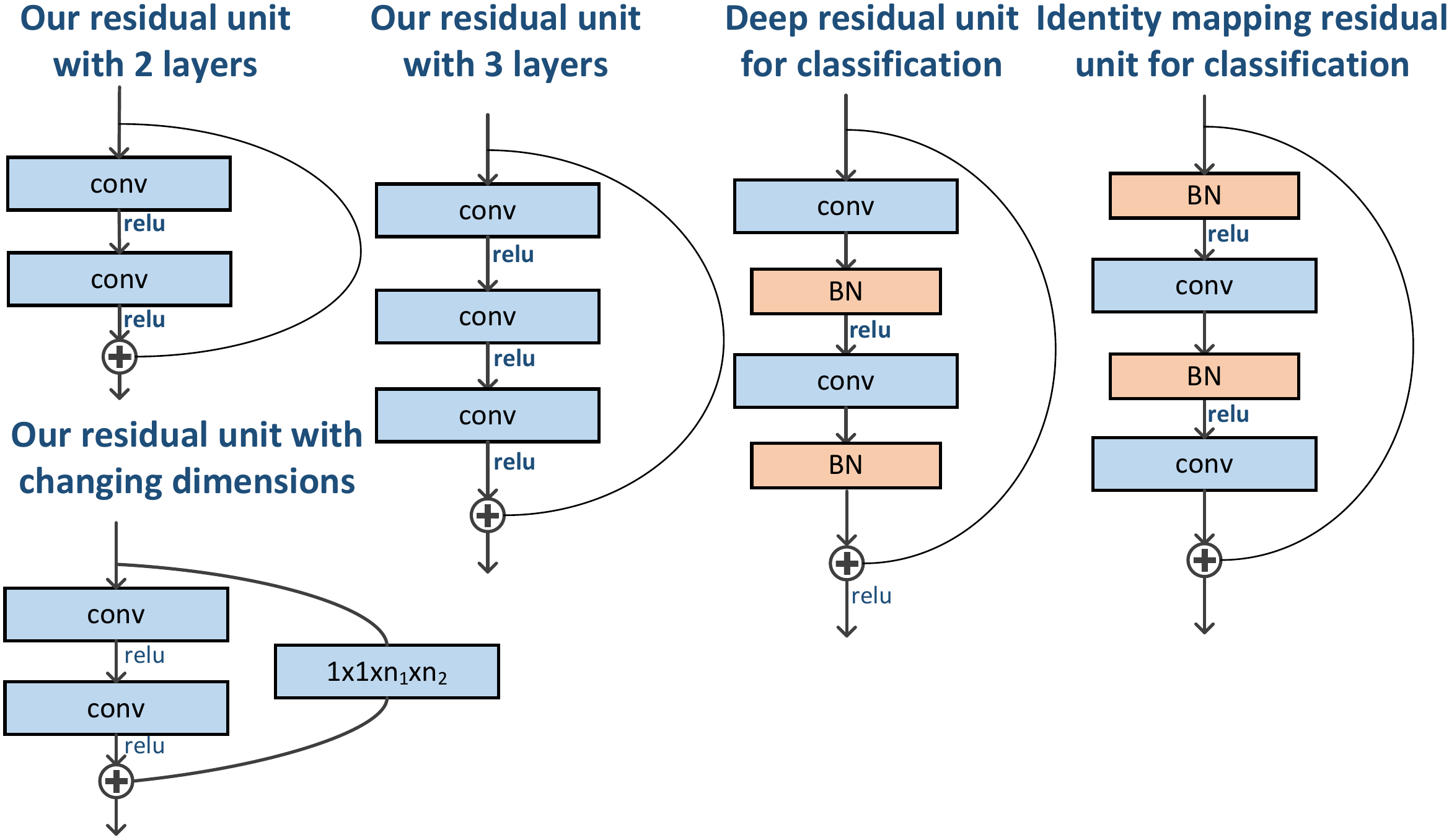}
     \caption{The architectures of different residual units.}
  \label{fig:res-unit}
  \end{figure*}

\subsection{Deep Residual Models}

The gradient exploding/vanishing problems are largely alleviated by skip connections in the deep residual models~\cite{he2015deep,he2016identity}. The architectures of our deep residual models especially the residual branches will be further designed from the perspective of centering activations.

Simply stacking the convolutional layers and rectified linear units as VDSR fashions~\cite{kim2016accurate} will have a mean activation larger than zero~\cite{clevert2015fast}. The non-zero mean activation acts as bias for the next layer. The more the layers are correlated, the higher their bias shift.

For a multi-layer perceptron network, Raiko \etal~\cite{raiko2012deep} proved that the transformation by shortcuts centered the activations which made the Fisher information matrix closer to a diagonal matrix, and thus standard gradient closer to the natural gradient. The transformations can be as
\begin{equation}\label{eq:transform}
                  x^{k+1}=\mathbf{A}\cdot \mathrm{T}(\mathbf{B}\cdot x^k)+\mathbf{C}\cdot x^k,
             \end{equation}
where $\mathbf{A}$, $\mathbf{B}$, $\mathbf{C}$ is the weight matrices, $\mathrm{T}$ is a nonlinearity activation, $x^k$ is the output of the neurons of the kth layer.

Similarly, for convolutional neural network, transformations can be as
\begin{equation}\label{eq:trans-conv}
                  x^{k+1}=f(\theta^k,x^k)+\mathbf{C}\cdot x^k,
             \end{equation}
where $f$ is a function composed by convolutions, nonlinearity activation, and Batch Normalization (BN). When the weight matrix $\mathbf{C}$ becomes identity matrix, function $f(\theta^k,x^k)$ will become our residual branches. Thus, our residual networks with skip connections can naturally centering activations and speed up learning.

For image super-resolution problems, super-resolution is only applied on the luminance channel (Y channel in YCbCr color space) in most of previous study~\cite{timofte2013anchored,dong2014learning,kim2016accurate}. It is obvious that the distribution of values on the luminance channel in the output HR images doesn't center at zero, while the residual images (high frequency of the images) have means towards zero. To center the activations, our deep residual CNN applies a large skip-connection as \cite{kim2016accurate} which makes the network predict the residual images (the high frequency of the images). Predicting the residual images has largely improved the training speed and convergency results.

Our deep residual CNN for image super-resolution is an end-to-end mapping model which can be roughly divided into three sub-networks to perform three steps: feature representation, nonlinear mapping, and reconstruction.

The feature representation sub-network extracts LR discriminative features from the LR input images, while nonlinear mapping part maps the LR feature representations into HR feature representations. Reconstruction part restores the HR images from HR feature representations. Feature representation sub-network applies plain network stacking convolutional and ReLU layers as shown in Fig.~\ref{fig:arch} and reconstruction sub-network only uses convolutional layers as~\cite{liang2017single}. The main body of our model, nonlinear mapping part consists of residual units which center the activations with shortcuts and ease the difficulties of training.

Typical units of our deep residual CNN are shown in Fig.~\ref{fig:res-unit}. Empirically, residual unit with 2 or 3 convolutional layers works well for image super-resolution problem, those two kinds of units are applied in the experiments. When featuremap dimensions change, the identity shortcut becomes a projection to change feature dimensions. The second right and rightmost are one unit of residual net for image classification problems proposed by He~\etal~\cite{he2015deep,he2016identity} respectively. Compared with them, the architectures of our residual functions are composed of convolutional, ReLU layers and shortcuts, which are quite different. Batch Normalization units are discarded and deployments are different. Batch Normalization~\cite{ioffe2015batch} reduces the distribution variations of layers (internal covariate) by normalizing the input of the layers. With an input x, the output of BN layer is given by
           \begin{equation}\label{eq:res-f}
                     BN_{\gamma,\beta}=\gamma (\frac{x-\mu}{\sqrt{\sigma^2+\varepsilon}})+\beta,
             \end{equation}
where $\gamma$ and $\beta$ are learnable parameters,  $\mu$ and $\sigma$ are the mean and variance of activations in the mini-batch, respectively, $\varepsilon$ is a small constant for numerical stability. Obviously, the activation after Batch Normalization operation has also been centered.  As skip connections (Eq.~\eqref{eq:trans-conv}) have naturally corrected the bias shift, thus if the residual network is not that deep\footnote{The bias from zero will accumulate as the network goes deeper.}, the BN layers can be abandoned as it needs extra learning and inference computations which take much more computational resources.

Shortcuts or skip connections which are identity mappings are realized by element-wise additions. As this element-wise addition increases very little computations, our feed-forward deep residual CNN has a similar computational complexity with VDSR~\cite{kim2016accurate} fashions network. Similar with {VDSR}~\cite{kim2016accurate}, small convolutional filter of size $3\times3$ has been applied. Assuming the input of $k$-th residual unit as $x^k$, the residual functions have the following form
             \begin{equation}\label{eq:res-f}
                     x^{k+1}=x^k+f(\theta^k,x^k),
             \end{equation}
where $\theta^k$ are the parameters of $k$-th residual unit.

A simple Euclidean loss function is adopted to make predictions approximate the high frequencies of examples
  \begin{equation}\label{eq:hf}
                 \mathcal{L}=\frac{1}{2n}\sum_{i=1}^n\|\mathcal{F}(\theta,I_i^l)-(I_i^h-I_i^l)\|^2
             \end{equation}
where $n$ is the number of patch pairs $(I^l, I^h)$, $F(\theta,I^l)$ denotes the predictions of our deep residual CNN with parameter $\theta$. Our deep residual CNN is composed of several \textbf{Container}s which have certain number of residual units. For succinctness, the filter numbers keep the same in each single container. The architectures of our deep residual CNN will be described as a sequence of the filter numbers (${N1}_{k1}$, ${N2}_{k2}$, $\cdots$) in containers. If subscript $k$ exists for $N_k$, it means there are $k$ residual units with each having a filter number of $N$ in this container.

   Stochastic gradient descent (SGD) with the standard back-propagation \cite{krizhevsky2012imagenet} is applied to train our deep residual CNN. In particular, the parameter is updated as Eq.~\eqref{eq:sto_Grad}, where $m$ denotes the momentum parameter with a value of 0.9 and $\eta$ is the learning rate.
                \begin{equation}
                \label{eq:sto_Grad}
                   \triangle_{i+1}= m\cdot\triangle_{i}+ \eta\cdot\frac{\partial loss}{\partial \theta_{i}},\quad \theta_{i+1}= \theta_{i}+\triangle_{i+1}
                \end{equation}

 High learning rates are expected to boost training with faster and better convergency. Adjustable gradient clipping \cite{kim2016accurate} is utilized to keep learning rates high while at the same time to prevent the net from gradient exploding problems. Gradients $\frac{\partial Loss}{\partial \theta_{i}}$ are clipped into the range of $[-\frac{\tau}{\eta},\frac{\tau}{\eta}]$, where $\tau$ is a constant value.

\subsection{Lightweight Design for the Proposed Model}

In this section, the `shape' of deep CNN has been explored to achieve better performances but with less number of parameters. The `shape' of deep CNN is determined by all the sizes and numbers of filters in each layer besides the depth of the network. Thin but small filter size works well with padding which leads to larger receptive field as network goes deep, in specific, $3\times3$ filter size has been applied. It is general that deeper and wider network will have larger model capacity and better feature representational ability. However, the number of parameters is restricted by the hardware or computational resources. Using less parameters to achieve better performances is essential for applications. Next, filter numbers and the combinations of filter numbers will be discussed to retrench parameters for a better performance.

\begin{figure*}[bhtp]
     \centering
     \footnotesize
     \includegraphics[width=0.800\textwidth]{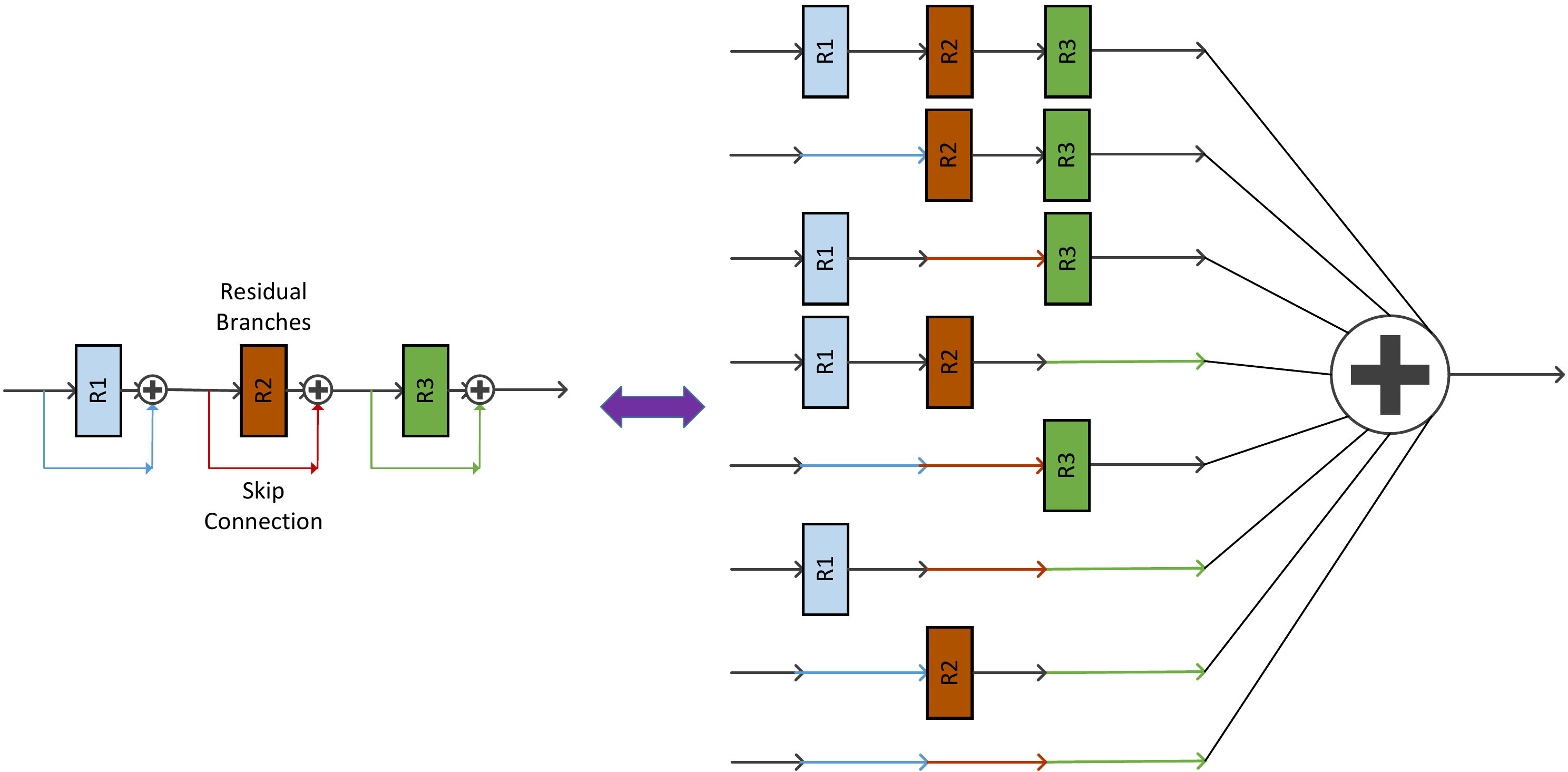}
     \caption{Residual network behaves like an ensemble of networks}
  \label{fig:ensemble}
\end{figure*}

\subsubsection{Exploring the Shape of the Architecture}

Inspired by the evolvement of Inception models~\cite{szegedy2015going,ioffe2015batch,szegedy2016rethinking,szegedy2017inception} and the bottle-neck architecture \cite{he2016identity}, it is supposed that changing the shape of the architecture may maintain the performance while largely reduce the computational parameters. Instead of applying $1\times1$ convolutions as bottle-neck architecture, the $3\times3$ convolutions are applied as image SR process largely depends on the contextual information in local neighbor areas.

The filter numbers of VDSR are kept the same. There seems to be few principles to decide filter numbers and the combinations of filter numbers in a network. Instead of using a same number of filters in a network, the filter numbers can be varied to potentially reduce parameters which could enable a deeper or wider network. 

Residual networks can be interpreted as an ensemble of many paths of differing depth~\cite{Veit2016Residual} and residual networks enable very deep networks by leveraging only the short paths during training~\cite{Veit2016Residual}. According to this assumption, if the models of short paths in the residual network have been less disturbed, the performance of residual network which is an ensemble could keep stable.

A strategy of gradually varying the `shapes' of residual models is proposed by us to reduce parameters. Gradually varying the shape of network means the filter numbers of the adjacent layers should increase or decrease gradually. This has been illustrated as Fig.~\ref{fig:ensemble}. In Fig.~\ref{fig:ensemble}, different residual branches and corresponding skip connections are denoted by different colors. The residual networks can be unfolded as a summations of models from different paths of residual networks. Considering a residual network with three units or sub residual network, if the filter numbers of the adjacent layers change gradually, \eg, only the filter numbers of R3 changes (\eg, decreases), a lot of paths are unaffected. Thus, the residual networks are more robust to the shape varying and our strategy can be applied to achieve better performances with less parameters.

\begin{figure*}[htb]
    \centering
     \setlength{\tabcolsep}{2pt}
  \footnotesize
  \begin{tabular}{ccccc}

  \includegraphics[width=0.200\textwidth]{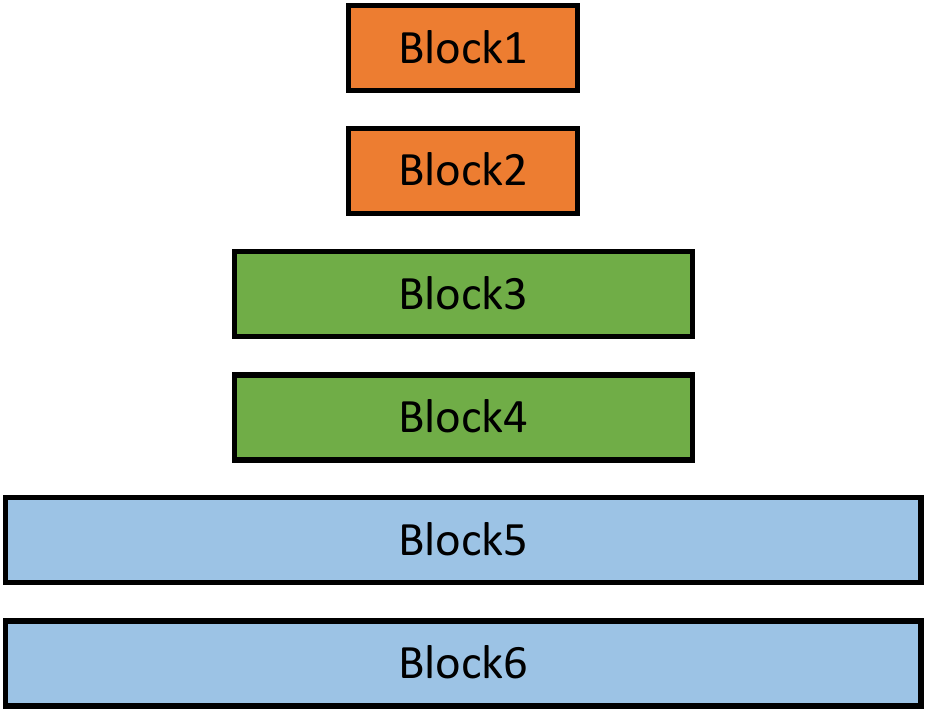} &
   \includegraphics[width=0.200\textwidth]{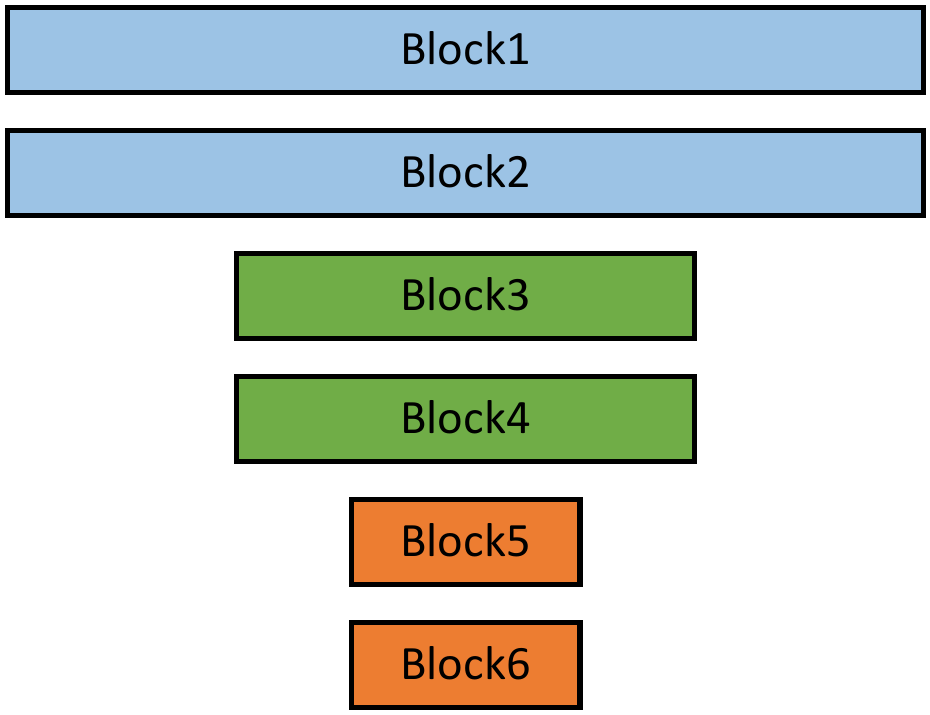} &
   \includegraphics[width=0.200\textwidth]{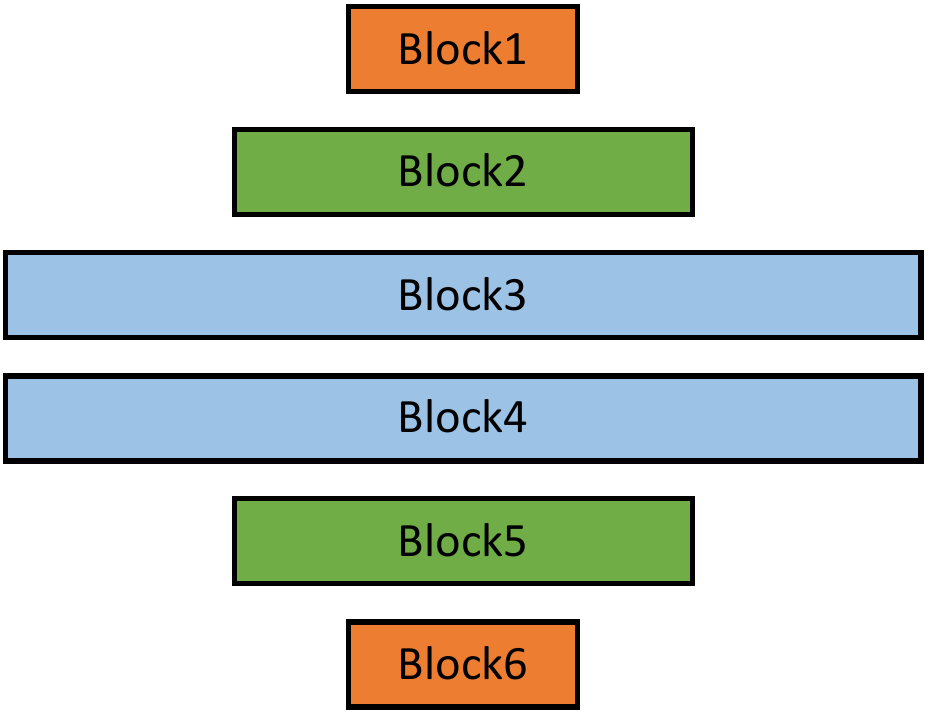} &
   \includegraphics[width=0.200\textwidth]{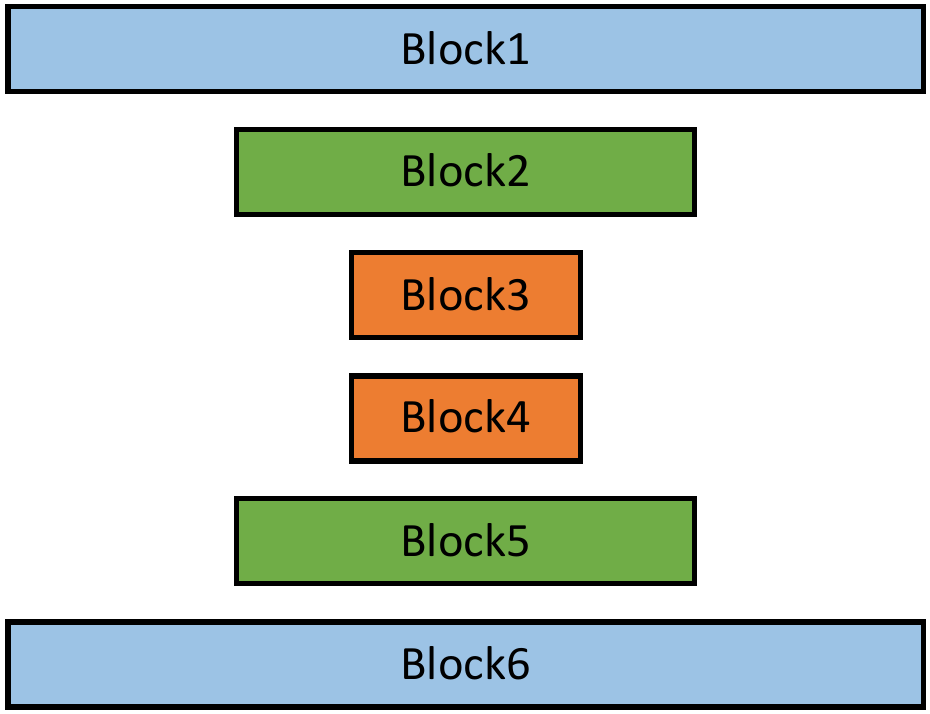} &
   \includegraphics[width=0.200\textwidth]{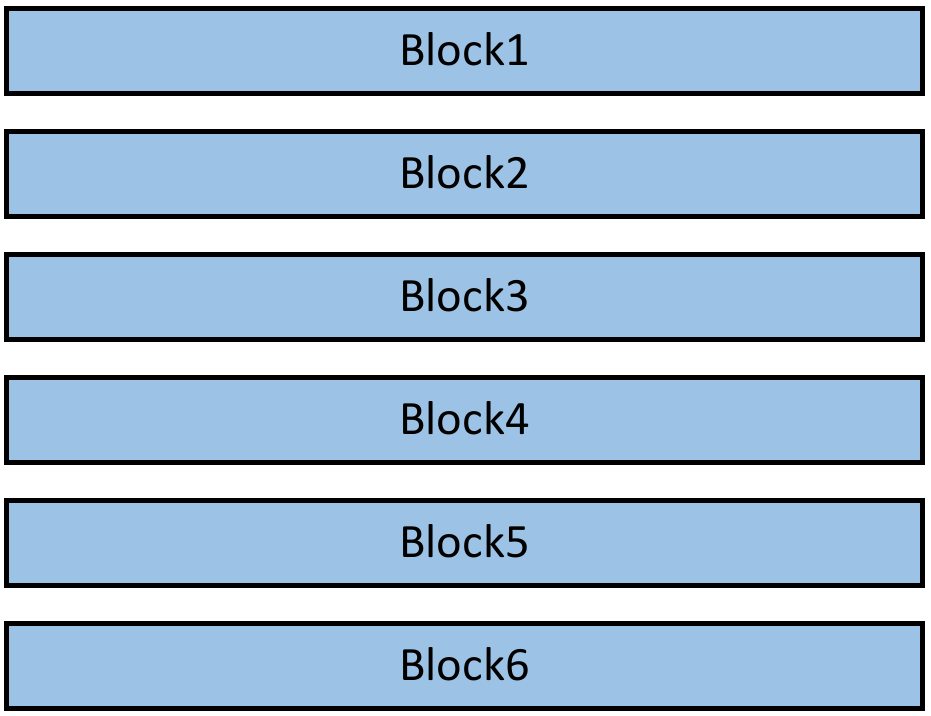} \\
   (a)&(b)&(c)&(d)&(e)\\
  \end{tabular}
  \caption{Different `shapes' of networks which gradually vary the feature map numbers. The width of the block correlates to the number of feature maps in the layer.}
  \label{fig:shapes}
\end{figure*}

The impacts of the feature map numbers in each layer on performance are carefully explored in the following fashions as Fig.~\ref{fig:shapes}: gradually, the numbers of feature maps 
 \begin{itemize}
   \item increase monotonically up to N.
   \item decrease monotonically from N.
   \item  increase up to N then decrease.
   \item decrease from N then increase.
   \item keep the same as N (baseline).
 \end{itemize}
In Fig.~\ref{fig:shapes}, the width of the square block correlates to the numbers of the feature map in the layer. The larger width of square block indicates there are more feature maps in that layer. Compared with the baseline way that the feature map numbers keep the same, applying gradually varying the shape strategy has largely reduced the parameters.

The experiments demonstrate that different lightweight designs have achieved comparable performances with less parameters. This will be further discussed in the experiments part. In comparison with our residual CNN, the performances of {VDSR} with different shapes fluctuate heavily. This proves our residual architectures are more robust to the shape varying of CNN and our strategy of gradually varying the `shape' of residual network could be applied to achieve better performances with less parameters.

 \subsubsection{Training with Multiple Upscaling Factors to Retrench Parameters}
 It has been pointed out that it is feasible to train a deep CNN for different upscaling factors~\cite{kim2016accurate}. Training datasets for different specified upscaling factors are combined together to enable our deep residual CNN to handle multiple upscaling factors, as images across different scales share some common structures and textures. Parameters are shared across different predefined upscaling factors which further dispenses with the trouble of retaining different models for different upscaling factors. It will retrench parameters when multiple upscaling factors are required.

   \begin{table*}[thb]
        \centering
         \caption{Comparison in different datasets and with different scales.}
         \label{table:results}
         \begin{tabular}{c|c|c|c|c|c|c|c|c|c}
         \hline
         \multirow{2}{*}{Dataset} & \multirow{2}{*}{Scale} & Bicubic & A+\cite{timofte2014a+} & RFL\cite{schulter2015fast} & SelfEx\cite{huang2015single} & SRCNN\cite{dong2014learning} & VDSR\cite{kim2016accurate} & SRResNetNB&R-basic\\
         && PSNR/SSIM & PSNR/SSIM & PSNR/SSIM & PSNR/SSIM & PSNR/SSIM & PSNR/SSIM & PSNR/SSIM& PSNR/SSIM\\
         \hline
         \multirow{3}{*}{Set5} & $\times$2 & 33.66/0.9299 & 36.54/0.9544 & 36.54/0.9537 & 36.49/0.9537 & 36.66/0.9542 & \textbf{37.53}/\textbf{0.9587} &37.51/\textbf{0.9587}&  37.27/0.9577\\
         & $\times$3 & 30.39/0.8682 & 32.58/0.9088 & 32.43/0.9057 & 32.58/0.9093 & 32.75/0.9090 & 33.66/0.9213 &\textbf{33.72}/\textbf{0.9215}& 33.43/0.9190\\
         & $\times$4 & 28.42/0.8104 & 30.28/0.8603 & 30.14/0.8548 & 30.31/0.8619 & 30.48/0.8628 & 31.35/\textbf{0.8838} & \textbf{31.37}/\textbf{0.8838}& 31.15/0.8796\\
         \hline
         \multirow{3}{*}{Set14} & $\times$2 & 30.24/0.8688 & 32.28/0.9056 & 32.26/0.9040 & 32.22/0.9034 & 32.42/0.9063 & 33.03/0.9124 & \textbf{33.10}/\textbf{0.9131}&32.86/0.9113  \\
         & $\times$3 & 27.55/0.7742 & 29.13/0.8188 & 29.05/0.8164 & 29.16/0.8196 & 29.28/0.8209 & 29.77/0.8314 & \textbf{29.80}/\textbf{0.8317} & 29.67/0.8297 \\
         & $\times$4 & 26.00/0.7027 & 27.32/0.7491 & 27.24/0.7451 & 27.40/0.7518 & 27.49/0.7503 & 28.01/0.7674 & \textbf{28.06}/\textbf{0.7681} &27.90/0.7648\\
         \hline
         \multirow{3}{*}{BSD100} & $\times$2 & 29.56/0.8431 & 31.21/0.8863 & 31.16/0.8840 & 31.18/0.8855 & 31.36/0.8879 & 31.90/0.8960 & \textbf{31.91}/\textbf{0.8961}&31.76/0.8940 \\
         &$\times$3 & 27.21/0.7385 & 28.29/0.7835 & 28.22/0.7806 & 28.29/0.7840 & 28.41/0.7863 & 28.82/0.7976 & \textbf{28.83}/\textbf{0.7980} &  28.73/0.7954\\
         &$\times$4 & 25.96/0.6675 & 26.82/0.7087 & 26.75/0.7054 & 26.84/0.7106 & 26.90/0.7101 & \textbf{27.29}/\textbf{0.7251} & 27.27/0.7248 &27.19/0.7221 \\
         \hline
         \multirow{3}{*}{Urban100} & $\times$2 & 26.88/0.8403 & 29.20/0.8938 & 29.11/0.8904 & 29.54/0.8967 & 29.50/0.8946 & 30.76/0.9140 & \textbf{30.88}/\textbf{0.9150}&30.47/0.9100   \\
         & $\times$3 & 24.46/0.7349 & 26.03/0.7973 & 25.86/0.7900 & 26.44/0.8088 & 26.24/0.7989 & 27.14/0.8279 & \textbf{27.17}/\textbf{0.8283}& 26.92/0.8208\\
         & $\times$4 & 23.14/0.6577 & 24.32/0.7183 & 24.19/0.7096 & 24.79/0.7374 & 24.52/0.7221 & 25.18/0.7524 & \textbf{25.22}/\textbf{0.7537}&  25.02/0.7452\\
         \hline
         \end{tabular}
   \end{table*}

\section{Experiments}

In this section, we conducted a series of experiments to explore the empirical principles to design a deep architecture for image super-resolution problem. The performances of the proposed method against the state-of-the-art SISR methods are compared which clearly demonstrate better or comparable subjective scores and more visual pleasing results.

The same 291 training images applied by VDSR were utilized for training, including 91 images proposed in Yang \etal \cite{yang2008image} and 200 natural images from Berkeley Segmentation Dataset (BSD). For testing, four datasets were investigated: `Set5' and `Set14' \cite{timofte2013anchored,dong2014learning},`Urban100' \cite{huang2015single} and `BSD100' \cite{timofte2013anchored,yang2014singleBenchmark}.

The size of example was set as $41\times41$ and the batch size was chosen as 64. Momentum and weight decay parameters were fixed as $0.9$ and $0.0001$ respectively. Multi-scale training was applied in all of the following experiments. Weight initialization methods \cite{he2015deep,he2016identity} were applied with small modulations. Learning rate was initially set to 0.1 and then decreased by a factor of 10 every 30 epochs. All these settings ensure us to make a fair comparison with the competing approaches including VDSR method.

\subsection{Comparisons with the State-of-the-art Methods}

Table \ref{table:results} shows the quantitative comparisons with A+~\cite{timofte2014a+}, RFL~\cite{schulter2015fast}, SelfEx~\cite{huang2015single}, SRCNN~\cite{dong2014learning} and VDSR~\cite{kim2016accurate}. Visual results are also represented to give intuitive assessment. In Table~\ref{table:results}, two models of our deep residual CNN with different depth have been investigated, denoted as \textbf{R-basic} and \textbf{SRResNetNB} respectively.  The residual unit in R-basic and deeper and larger model SRResNetNB has two convolutional layers. R-basic $(16_3,32_3,64_3)$ has 22 layers, while SRResNetNB $(16_3,32_3,64_3,128_3,256_3)$ has 34 layers. SRResNetNB has achieved the best performances compared with other methods in most cases and comparable results in other situations.

\begin{figure}[!thb]
  \centering
  \footnotesize
    \includegraphics[width=0.40\textwidth]{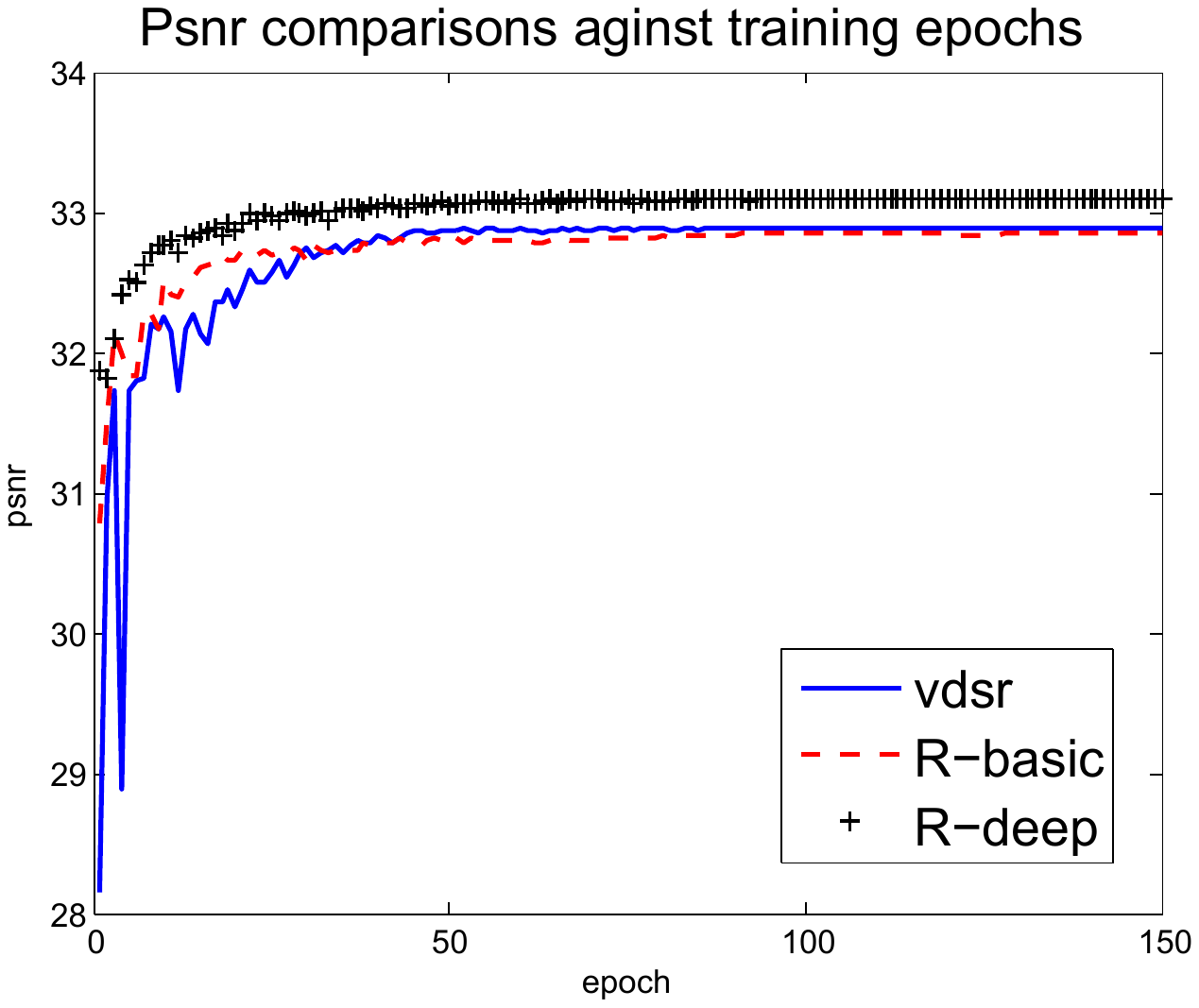}
  \caption{Comparisons of test psnr on Set 14 against training epochs among SRResNetNB(denoted as R-deep), R-basic and VDSR.}
  \label{fig:Comp_a}
\end{figure}

R-basic outperforms the other methods except VDSR. However, the performances of VDSR (20 layers)  have not been obtained by our reimplementation. For example, the average PSNR of VDSR by our reimplementation for Set5 and Set14 are 37.32dB and 32.89dB respectively, with a gap of more than 0.1db from the reported results. Assisted with the missing tricks, the performance of our model is expected to be further boosted. In Fig.~\ref{fig:Comp_a}, the PSNR against training epochs has been compared among R-basic, SRResNetNB, and VDSR trained by us. Deeper and larger model SRResNetNB outperformed {VDSR} at very beginning with a large margin. Although R-basic contains much less parameters, R-basic model has obtained comparable performances with VDSR.

In Fig.~\ref{fig:SRresults}, all the compared results are obtained by the released code of the authors or from the reported ones in the paper. Visually pleasing results have been achieved by our model. Restorations of our method contain more authentic texture and more clear details compared with the results by other methods such as the texture of the zebra head. Our method has provided less artifacts, \eg, all the other methods except ours have restored obvious artifacts at the location of book. Shaper edges have appeared in our restorations which have represented visually more pleasing results.

\subsection{Number of Parameters}

For R-basic model, there are 22 convolutional layers and 0.3M(322721) parameters accumulated by the numbers of corresponding weights and bias. For SRResNetNB model, 34 convolutional layers and 5M(4975905) parameters are applied. The compared VDSR in Table \ref{table:results} is 20 layers and has 0.7M(664704) parameters.  Although SRResNetNB has more parameters, our SRResNetNB model is still acceptable which can be efficiently trained with a single GPU.

\subsection{The Position of RelU}

\begin{figure}[thb]
  \centering
  \footnotesize
  \begin{tabular}{cc}

  \subfigure[ReLU before convolution]{\includegraphics[width=0.220\textwidth]{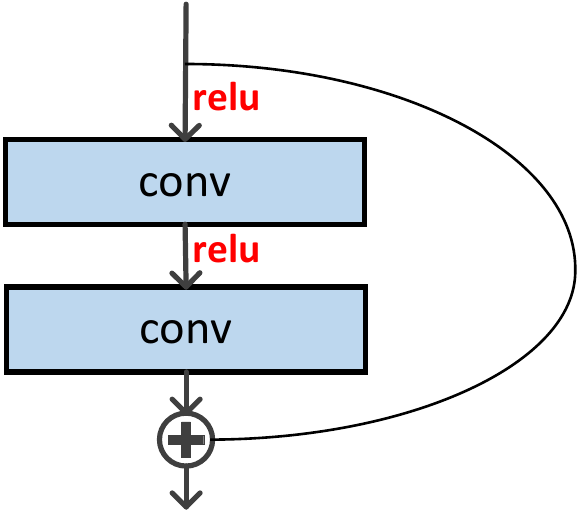}} &
    \subfigure[ReLU after convolution]{\includegraphics[width=0.220\textwidth]{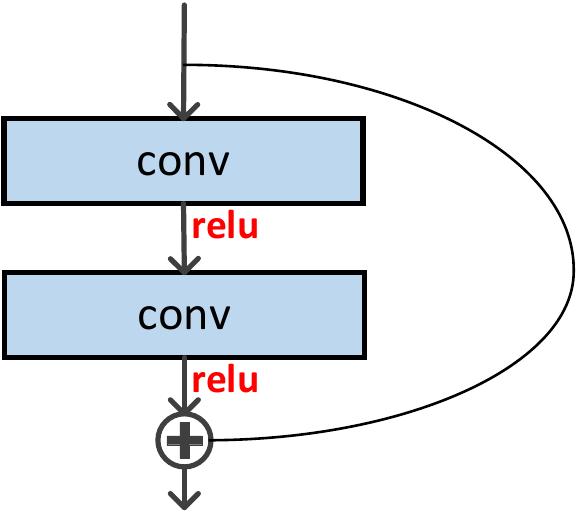}} \\
  \end{tabular}
  \caption{The positions of ReLU in residual branches.}
  \label{fig:mmm_Comp}
\end{figure}

In the residual branches, convolutional and ReLU layers are applied. The performances compared with the positions of ReLU layers (ReLU before/after conv) as in Fig.~\ref{fig:mmm_Comp} are represented in Table~\ref{table:mmm_ablations} on Set14. The compared network has a same depth and corresponding convolutional layers among these networks have the same parameter numbers.

\begin{table}[thb]
  \centering
  \caption{Ablation comparisons for residual network with different orders of convolution and ReLU layers in terms of average PSNR (dB) on Set14.}
  \label{table:mmm_ablations}
  \begin{tabular}{c|c|c}
    \hline
      scale &identity$+$&identity$+$\\
		   &ReLU after conv&ReLU before conv\\
    \hline
      $\times$ 2 &   32.97    & 33.01\\
      $\times$ 3 &     29.75   & 29.77  \\
      $\times$ 4 &    28.02   & 28.02  \\
    \hline
 \end{tabular}
\end{table}

From the results in Table~\ref{table:mmm_ablations}, we conclude that the positions of ReLU in the residual branches make small differences.

\subsection{Impacts of Batch Normalization on SISR}

\begin{figure}[!hbt]
  \centering
  \footnotesize
    \includegraphics[width=0.40\textwidth]{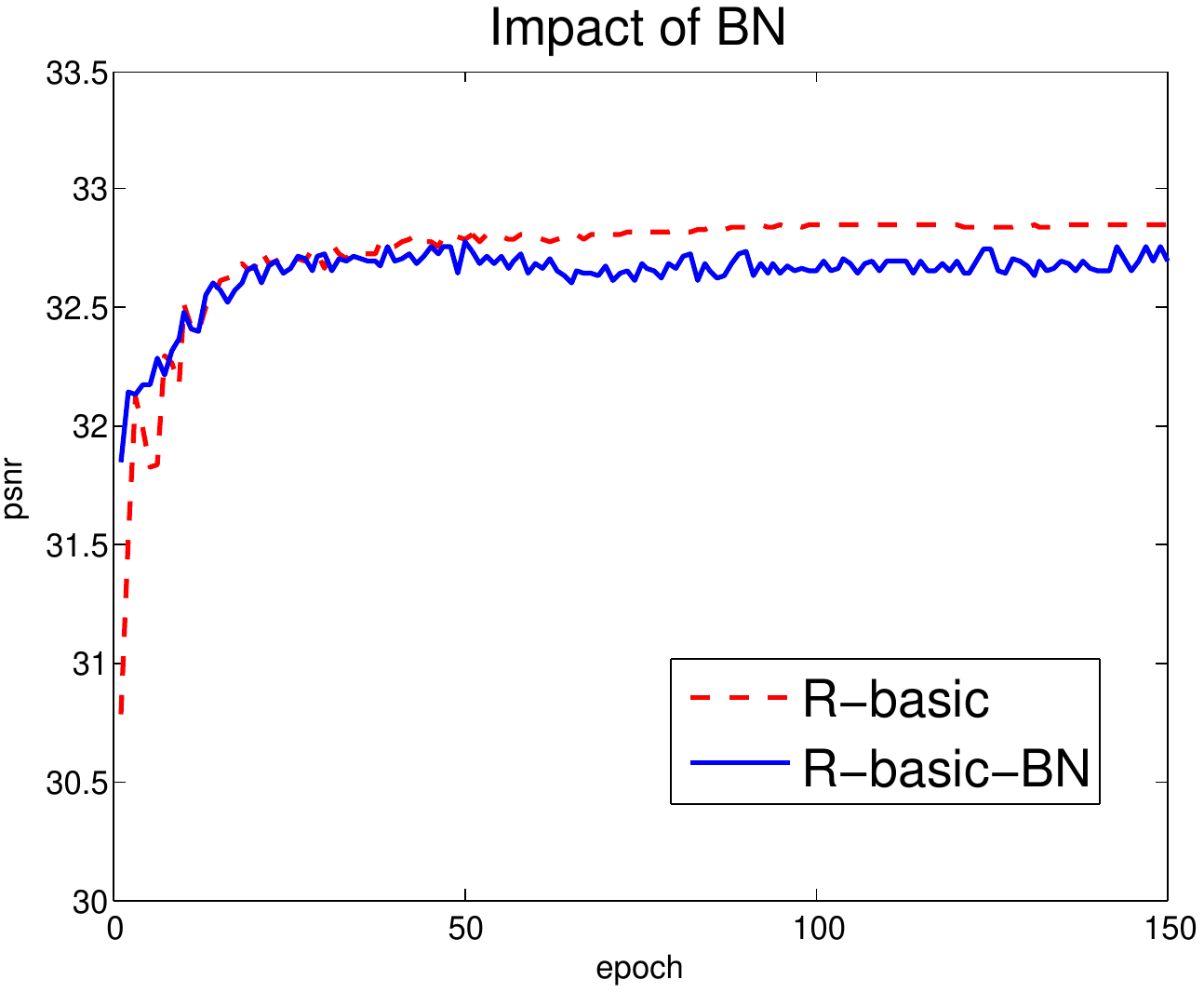}
  \caption{The impacts of BN: test psnr of Set 14 against training epochs by our R-basic with and without BN.}
  \label{fig:Comp_b}
\end{figure}

In Fig.~\ref{fig:Comp_b}, test PSNR of Set 14 against training epochs by our R-basic with and without BN are compared to demonstrate the impacts of Batch Normalization on SISR problems.
In Fig.~\ref{fig:Comp_b}, the compared structure with BN layers is the same as the structure applied for image classifications~\cite{he2016identity}, showed in the rightmost column in Fig.~\ref{fig:res-unit}. 

It seems adding BN operations has hampered further improvement when more epoches have been performed. Normalizing input distribution of mini-batch to suppress data shifting has been proved powerful and largely accelerated the training convergency speed. It also enables deeper architectures and larger learning rates to be utilized in other tasks. However, whiten input and output of the intermediate layer may not be suitable for image super-resolution task which needs precise output. Another suspect may be regularization effects of BN have not been fully exploited as the training set of Fig.~\ref{fig:Comp_b} is still limited in contrast with ImageNet. As larger learning rates were enabled by gradient clipping methods, the benefits of BN for leaning rates are alleviated.

From the perspective of centering activations, the skip connection itself has the benefits of centering the activations which partially reduces the necessities of BN operations when the network is not too deep to correct the mean bias. Moreover, the BN operation takes extra computations during learning and inference. Without BN operation, provided with certain computational resources, larger and wider deep architectures can be enabled to get better performances.

The impacts of Batch Normalization on SISR are still an open issue for the future study.

\subsection{The Deeper the Better, the Wider the Better}

SRResNetNB performs much better than R-basic and VDSR model with deeper and wider network. Next, ablations of our system would be evaluated to unpack this performance gain. The skip connections and two factors, width (related to filter numbers) and depth of our model would be analyzed in the following steps.

\begin{table}[thb]
       \centering
            \caption{PSNR comparison between our residual CNN and {VDSR} trained by us}
           \label{table:VDSR-R}
            \begin{tabular}{c|c|c|c|c}
            \hline
            &Set5&Set14&BSD100&Urban100\\
             \hline
             \hline
            $R(64_8)$&37.28&32.91&31.72&30.45\\
            \hline
            VDSR&37.32&32.89&31.77&30.51\\
             \hline
            \end{tabular}
\end{table}

First, 20-layer VDSR has been added with 8 identity shortcuts to form a residual network, denoted $R(64_8)$. Each residual unit has two convolutional and Relu layers. The performance of $R(64_8)$ is roughly the same as VDSR in Table~\ref{table:VDSR-R}. The shortcuts have very little impacts on the descriptive power. From the perspective of the centering the activations, to predict the high frequencies of image has pushed the final output activation centered. Within certain depth (\eg,~20), the difficulties of learning has been alleviated, thus the shortcuts of the residual network have less impacts on the descriptive power.
\begin{table*}[!hbt]
       \centering
            \caption{PSNR by the residual model of different depth and width with a magnification factors 2 in Set14.}
           \label{table:dw}
            \begin{tabular}{c|c|c|c|c|c}
            \hline
           & $R(16_3,32_3,64_3)$ &$R(32_3,64_3,128_3)$ &$R(16_3,32_3,64_3,128_3)$& $R(16_3,32_3,64_3,128_3,256_3)$ &$R(4_3,8_3,16_3,32_3,64_3)$\\
           \hline
           \hline
           PSNR(dB)&32.85& 32.96 & 33.00 &33.10&32.91\\
             \hline
            \end{tabular}
\end{table*}

If the network goes even deeper, the mean bias accumulate and difficulties of training increase. Then the benefits of skip connection will dominate that it alleviates gradient vanishing/exploding problems and helps centering the activations in the layers of the net, which enable a deeper network and greatly improve the performance.

Second, fixing the depth of the model, simply broadening the width will improve the performance as showed in Table~\ref{table:dw}, \eg, $R(16_3,32_3,64_3)$ vs $R(32_3,64_3,128_3)$, $R(4_3,8_3,16_3,32_3,64_3)$ vs $R(16_3,32_3,64_3,128_3,256_3)$. Increasing the filter numbers would enlarge the model capacity which enables modeling more complex nonlinear mappings from LR examples to HR examples.

Third, the deeper the architecture, the better the performance. Adding more residual units, \eg, $R(16_3,32_3,64_3,128_3)$ vs $R(32_3,64_3,128_3)$ will improve the performance. Certainly, the depth should be no more than certain limit to avoid the overfitting problem and computational resource limitations. Within this limit, the deeper the better. Our residual unit eases the training difficulties which enables a deeper CNN architecture to improve the situation. On the other side, when model goes deeper as our residual SRResNetNB, plain deep CNN like VDSR fashions can not converge well and the restorations deteriorate. Another attempt to facilitate deeper net is the lightweight design which aims to solve the problem of too many parameters. It will be discussed next.

\subsection{Lightweight Design}

In this part, the proposed strategy of gradually varying the `shape' of residual network has been investigated.  The performances of different architectures with different shapes have been investigated for our residual net in Table~\ref{table:shape-R} and {VDSR} fashions in Table~\ref{table:shape-V} counterpart.

The number of featuremap has been gradually varied. To be specific, there are 28 layers as 6 \textbf{container}s stack, each \textbf{container} contains 2 residual units (2 convolutional layers in each residual unit). The depth can be calculated as $28=2+6\times2\times2+2$, where feature representation sub-network and reconstruction sub-network each have 2 convolutional layers. For models of VDSR fashions, 12-layer VDSR have been explored. For residual architectures, networks of different `shapes' have achieved comparable results. On the contrary, the performances of VDSR structures have largely fluctuated when the shapes of the networks vary. 
\begin{table*}[thb]
\centering
\caption{Performance by different residual models which have different shapes with a magnification factor 2 in Set14.}
\label{table:shape-R}
\scriptsize
\begin{tabular}{c|c|c|c|c}
            \hline
           residual& $R(16_4,32_4,64_4)$& $R(64_4,32_4,16_4)$ & $R(16_2,32_2,64_2,64_2,32_2,16_2)$& $R(64_2,32_2,16_2,16_2,32_2,64_2)$\\
            \hline
             \hline
            PSNR(dB)& 32.91&32.85&32.94&32.89\\
            \hline
\end{tabular}
\end{table*}

\begin{table*}[thb]
\centering
\caption{Performance by {VDSR} models which have different shapes with a magnification factor 2 in Set14.}
\label{table:shape-V}
\begin{tabular}{c|c|c|c|c|c}
            \hline
            {VDSR}& $(8_2,16_2,64_2)$ &$(64_2,16_2,8_2)$ & $(8,16,64,64,16,8)$&$(64,16,8,8,16,64)$& $(64_{16})$\\
            \hline
            \hline
              PSNR(dB)&32.68&32.59&32.66 &32.50&32.85\\
            \hline
\end{tabular}
\end{table*}

Residual networks can be interpreted as an ensemble of models which are the paths of differing depth in the residual network~\cite{Veit2016Residual}. When the `shapes' of residual models are gradually changing, some short paths in the residual network have been less disturbed as Fig.~\ref{fig:ensemble}. Thus, the performances of the ensembles are nearly unchanged.  On the contrary, the single path VDSR network are more disturbed by the variations of the shape. Instead of keeping the filter number fixed, less parameters can be applied for the residual network with our strategy to achieve comparable performances.

\subsection{Training with Multiple vs Single Upscaling Factors}

In this section, we compare the performances of networks handling multiple with respect to single upscaling factors as~\cite{kim2016accurate} in Table~\ref{table:Multi2single}. The training examples from different upscaling factors were mixed together to enable the model handling multiple upscaling factors. It seems mixing samples augmentations strategy~\cite{kim2016accurate} from different upscaling factors has slightly boosted the performances, especially for large upscaling factors.

\begin{table}[thb]
       \centering
       \setlength{\tabcolsep}{3.5pt}
            \caption{PSNR comparisons between models handling multiple vs single upscaling factors, denoted as `Multiscale' and `single scale'}
           \label{table:Multi2single}
            \begin{tabular}{c|c|c|c|c|c}
            \hline
                 \multicolumn{2}{c}{} &Set5&Set14&BSD100&Urban100\\
             \hline
                   \multicolumn{2}{c}{} & PSNR/SSIM & PSNR/SSIM & PSNR/SSIM & PSNR/SSIM \\
             \hline
           \multirow{2}{*}{$\times$ 2}& Multiscale& 37.51/0.9587&  \textbf{33.10}/\textbf{0.9131}&\textbf{31.91}/\textbf{0.8961} & \textbf{30.88}/\textbf{0.9150}\\

            &single scale& \textbf{37.52}/\textbf{0.9589}&33.03/0.9129&31.90/0.8958&30.84/0.9143\\
             \hline
             \hline
               \multirow{2}{*}{$\times$ 3}& Multiscale& \textbf{33.72}/\textbf{0.9215}& \textbf{29.80}/\textbf{0.8317} &\textbf{28.83}/\textbf{0.7980} & \textbf{27.17}/\textbf{0.8283}\\

            &single scale&33.6/0.9212 & 29.75/0.8313&28.79/0.7967& 27.08/0.8255\\
             \hline
              \hline
               \multirow{2}{*}{$\times$ 4}& Multiscale&\textbf{31.37}/\textbf{0.8838} &\textbf{28.06}/\textbf{0.7681}  &\textbf{27.29}/\textbf{0.7251} &\textbf{25.22}/\textbf{0.7537} \\

            &single scale&31.30/0.8824& 27.99/0.7668&27.24/0.7237& 25.14/0.7051\\
             \hline
            \end{tabular}
\end{table}

\section{conclusion}

In this paper, from the perspective of centering activations and ensemble behaviors of residual network, a novel residual deep CNN which takes advantage of skip connections or identity mapping shortcuts in avoiding gradient exploding/vanishing problem was proposed for single image super-resolution. In particular, the `shape' of CNN has been carefully designed such that a very deep convolutional neural network with much fewer parameters can produce even better performance. Based on the investigations into the influences of the network `shape' on the performances, a strategy of gradually varying the `shape' of the network has been proposed to construct this lightweight model.
Experimental results have demonstrated that the proposed method can not only achieve state-of-the-art PSNR and SSIM results for single image super-resolution but also produce visually pleasant results.

\begin{figure*}[th]
  \centering
  \setlength{\tabcolsep}{2pt}
  \footnotesize
  \begin{tabular}{p{1cm}cccc}
    $\bigotimes{2}$&
     \subfigure{\includegraphics[width=0.20\textwidth]{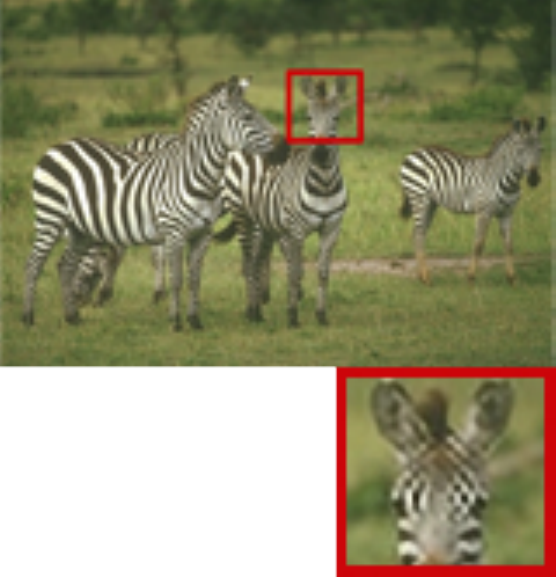}}&
    \subfigure{\includegraphics[width=0.20\textwidth]{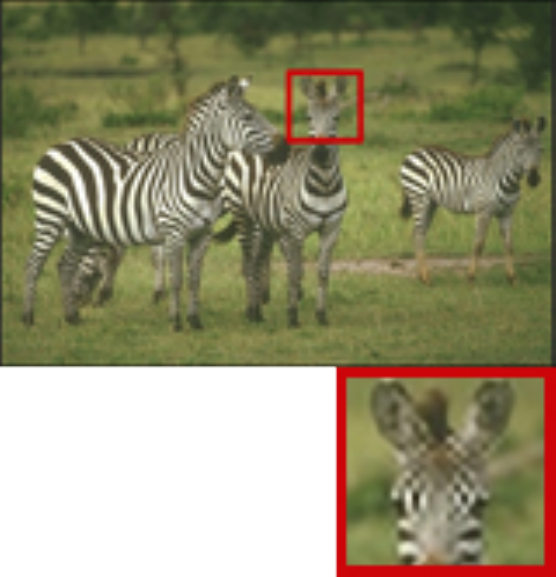}} &
    \subfigure{\includegraphics[width=0.20\textwidth]{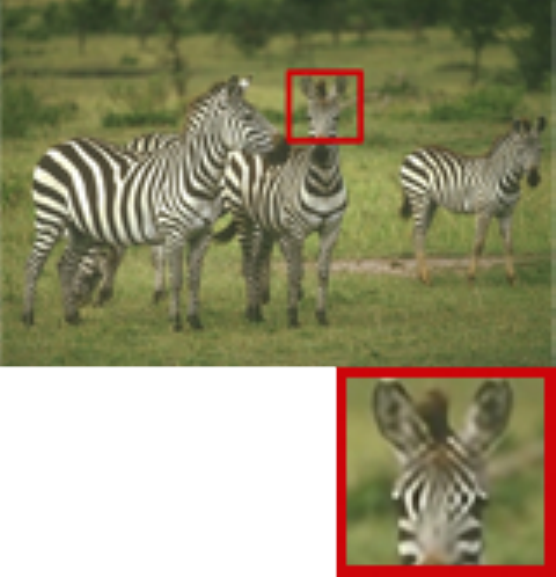}}&
      \subfigure{\includegraphics[width=0.20\textwidth]{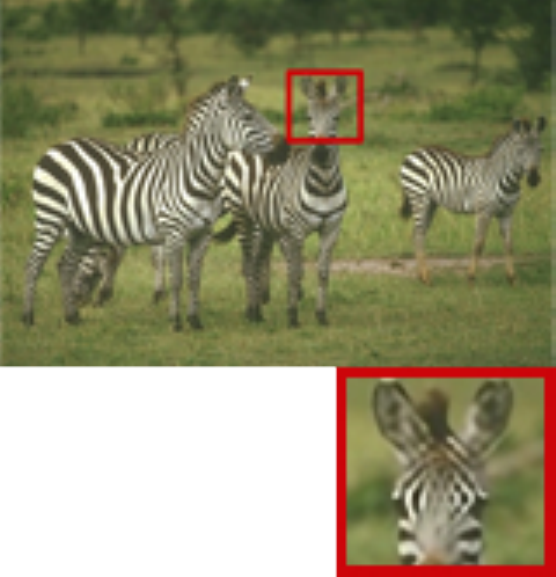}} \\
     &27.35db &27.24db& 28.61db&29.01db\\
       $\bigotimes{3}$ &
       \subfigure{\includegraphics[width=0.20\textwidth]{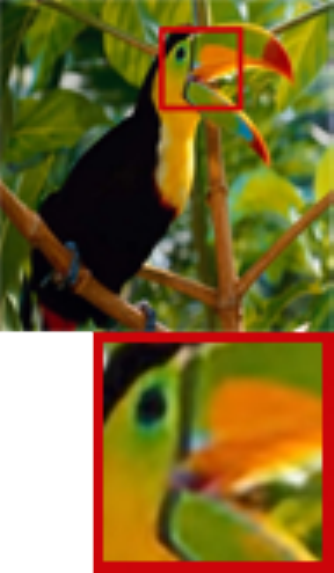}}&
    \subfigure{\includegraphics[width=0.20\textwidth]{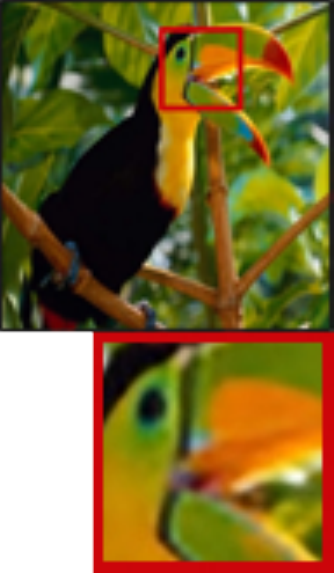}} &
    \subfigure{\includegraphics[width=0.20\textwidth]{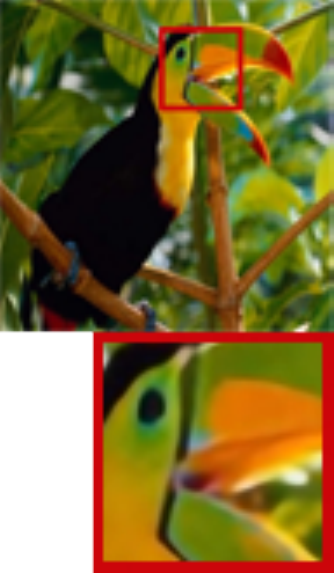}}&
      \subfigure{\includegraphics[width=0.20\textwidth]{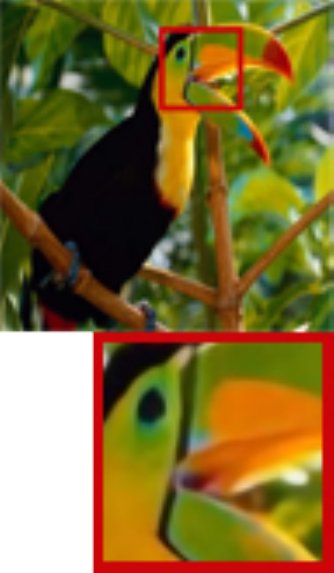}} \\
     &34.93db& 35.20db &  36.67db & 36.91db\\
       $\bigotimes{4}$ &
       \subfigure{\includegraphics[width=0.20\textwidth]{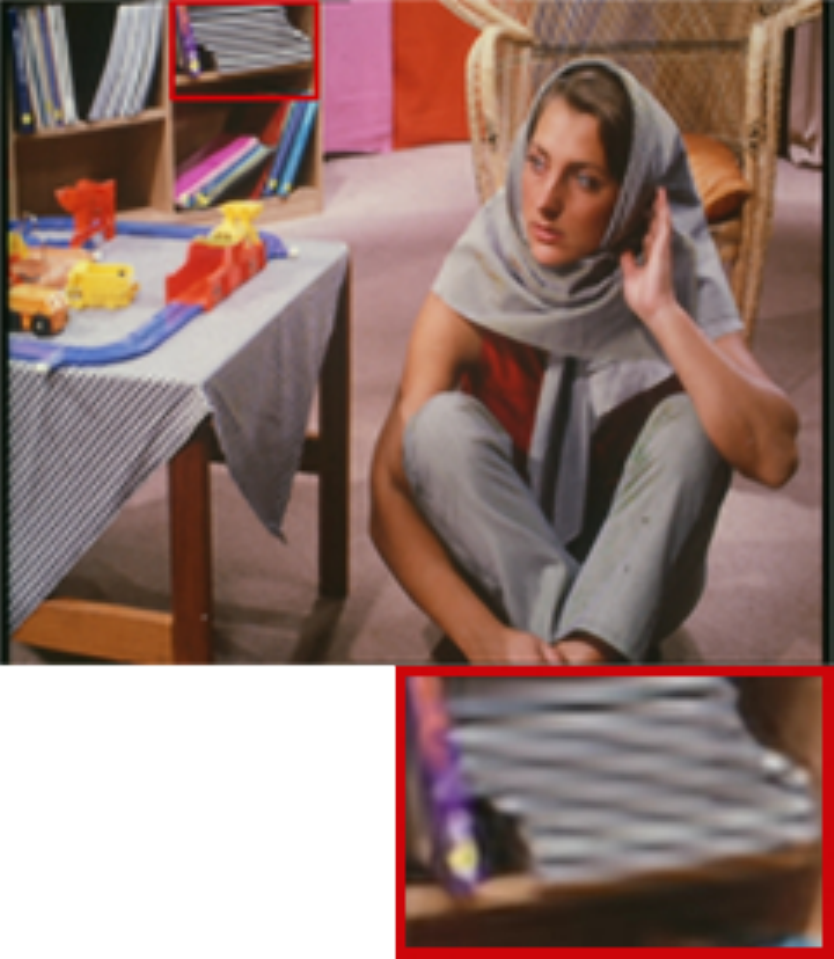}}&
    \subfigure{\includegraphics[width=0.20\textwidth]{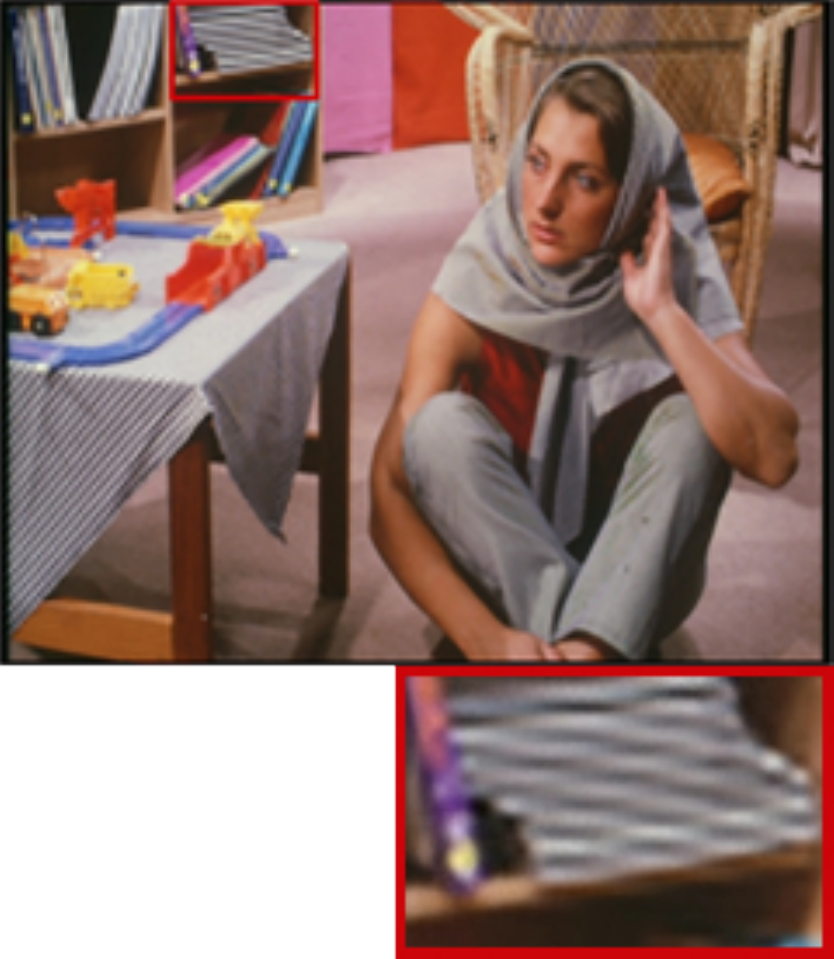}} &
    \subfigure{\includegraphics[width=0.20\textwidth]{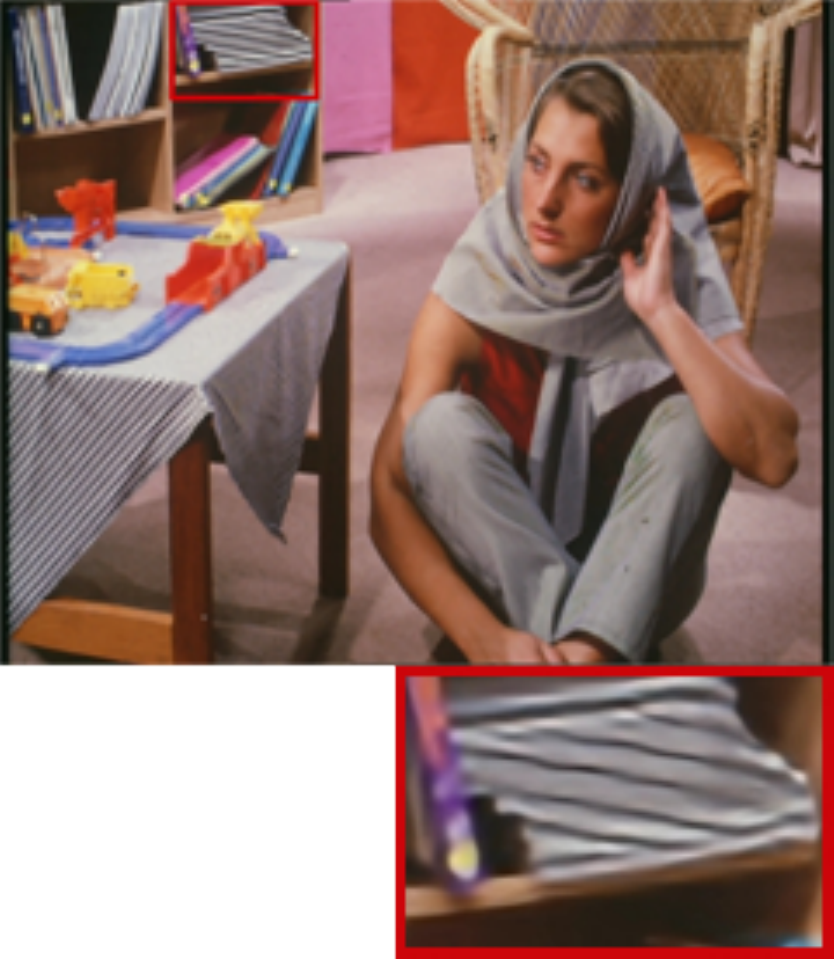}}&
      \subfigure{\includegraphics[width=0.20\textwidth]{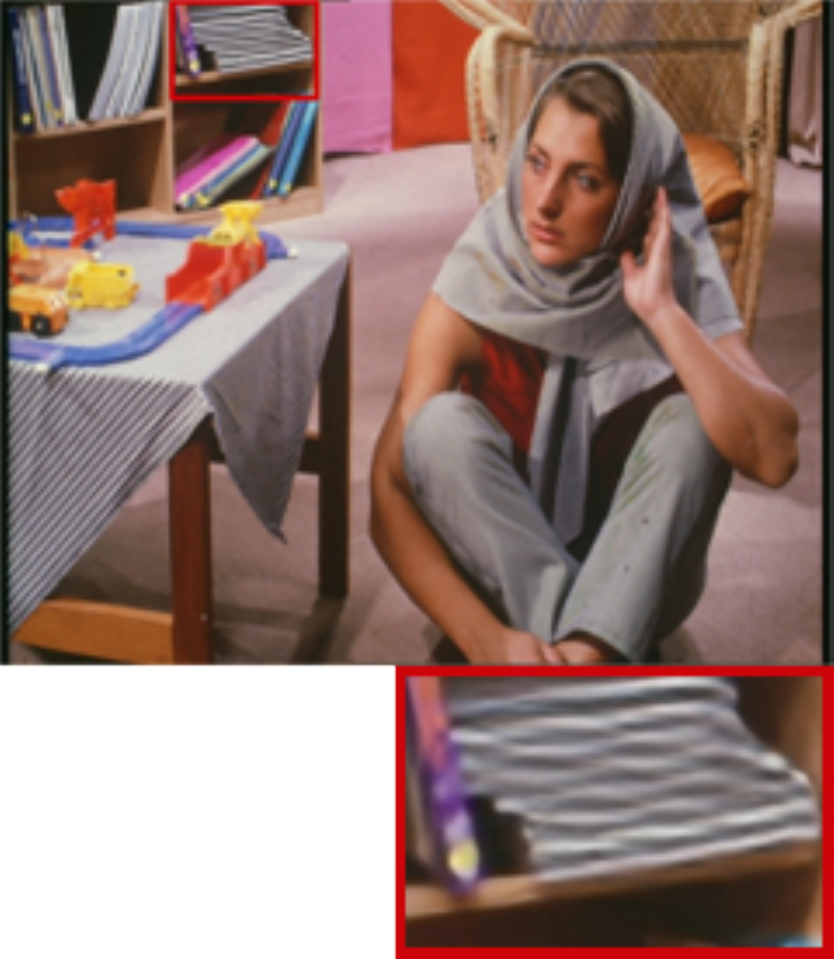}} \\
   & 25.70db & 25.72db &25.79db  &26.00db\\
    upscaling factors &\textbf{SRCNN}\cite{dong2014learning} & \textbf{RFL}\cite{schulter2015fast} & \textbf{VDSR}\cite{kim2016accurate} & \textbf{ours}\\
    GT&
    \subfigure{\includegraphics[width=0.20\textwidth]{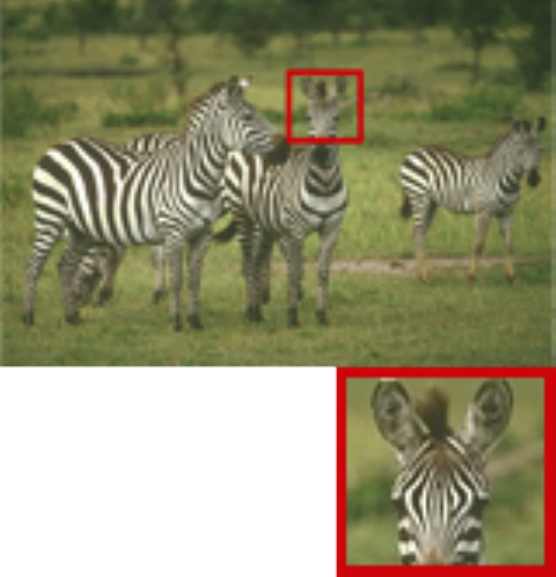}}&
    \subfigure{\includegraphics[width=0.20\textwidth]{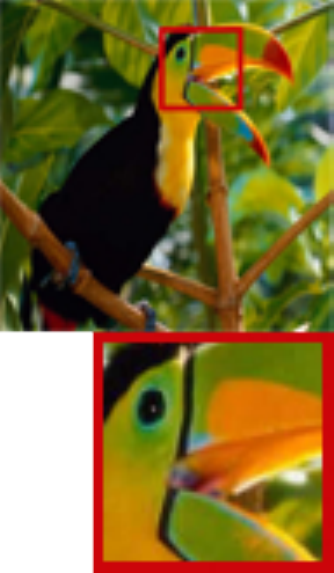}} &
    \subfigure{\includegraphics[width=0.20\textwidth]{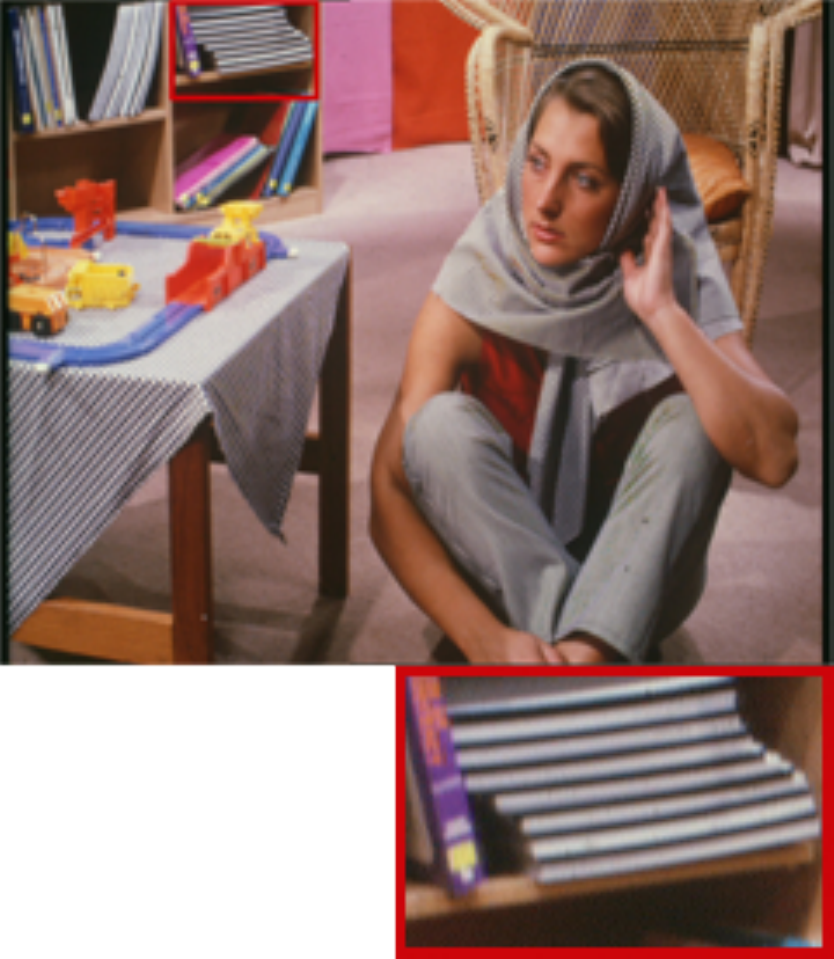}}&\\
  \end{tabular}
  \caption{Comparisons of image SR results with different methods in different upscaling factors}
  \label{fig:SRresults}
\end{figure*}



%



\section*{Acknowledgment}

 This work is partially supported by National Science Foundation of China under
Grant NO. 61473219.


{\small
\bibliographystyle{IEEEtran}
\bibliography{ImgSR}
}

\end{document}